\documentclass{article}

\usepackage{microtype}
\usepackage{graphicx}
\usepackage{subcaption}
\usepackage{booktabs}
\usepackage{multirow}
\usepackage{array}
\usepackage{enumitem}
\usepackage{amsmath}
\usepackage{amssymb}
\usepackage{mathtools}
\usepackage{hyperref}

\usepackage[preprint]{icml2026}
\usepackage[capitalize,noabbrev]{cleveref}
\icmltitlerunning{Risk-Sensitive Evaluation of LLM Contract Clause Generation}

\begin{document}

\twocolumn[
\icmltitle{Beyond Accuracy and Surface Fluency: Risk-Sensitive Evaluation of LLMs for Legal Clause Generation}

\begin{icmlauthorlist}
  \icmlauthor{Sundaraparipurnan Narayanan}{aite}
  \icmlauthor{Devansh Singh}{aite}
\end{icmlauthorlist}

\icmlaffiliation{aite}{AI Tech Ethics}

\icmlcorrespondingauthor{Sundaraparipurnan Narayanan}{sundar.narayanan@aitechethics.com}
\icmlcorrespondingauthor{Devansh Singh}{devansh.singh@aitechethics.com}

\icmlkeywords{Legal AI, Large Language Models, Contract Drafting, Evaluation, AI Governance}

\vskip 0.3in

]

\printAffiliationsAndNotice{}

\begingroup
\renewcommand\thefootnote{}
\footnotetext{Accepted to the AI for Law Workshop @ ICML 2026.}
\endgroup

\begin{abstract}
Large language models (LLMs) are increasingly used to draft contractual language, yet conventional accuracy or preference-based evaluations are poorly matched to legal drafting. A clause may be fluent and stylistically polished while still omitting an essential carve-out, allocating risk in an unenforceable way, assuming an inapplicable jurisdiction, or exposing a party to regulatory liability. This paper presents a empirical study design and framework for evaluating LLM-generated contract clauses. The study evaluates four models - Claude Haiku 4.5, Gemini 2.5 Flash Lite, GPT 5.4 Nano, and Qwen 3.5 Flash, across 22 contract clause categories and 34 legally-motivated failure modes. We combine two evaluation frameworks: CLAUSE, which classifies prompts by legal function and failure target, and LENS-CRAFT, which scores outputs across nine legal-quality dimensions. Instead of averaging dimension scores, the study applies a Max Severity Principle so that a single legally decisive defect remains visible. The paper provides the evaluation protocol, taxonomy, analysis plan, and a results structure for reporting empirical findings. We argue that legal AI evaluation should move beyond aggregate accuracy toward clause-specific, failure-mode-driven, and risk-sensitive assessment.
\end{abstract}

\section{Introduction}
LLMs have emerged to produce contract clauses that appear polished, complete, and professional. In legal practice, however, drafting quality is not reducible to lingustic fluency. A generated privacy clause may fail to distinguish controller and processor obligations; a limitation-of-liability clause may cap liability in circumstances where public policy, statute, or negotiation context requires exceptions; and a jurisdiction clause may assume an inappropriate forum. These errors are not merely legal representation, but may affect enforceability, compliance, create litigation exposure, limit negotiation leverage, and amplify professional responsibility.

Legal work imposes evidentiary expectations that differ from generic Natural Language Processing (NLP) tasks. Prior work shows that LLMs can hallucinate legal contexts, struggle with jurisdiction-specific questions, and underperform on basic legal text tasks \citep{dahl2024large,blair2024blt,hou2024gaps}. Legal benchmarks and legal-agent evaluations provide important coverage of legal reasoning, retrieval, and workflow tasks \citep{guha2023legalbench,pipitone2024legalbenchrag,li2024legalagentbench,wang2024legal}. However, contract drafting remains distinct as the output is not only an answer but also potential operative legal language (which requires expertise and jurisdictional context).

Our research studies the aforementioned use case through a structured evaluation mechanism for generating, stress-testing, and scoring contractual clauses. The protocol combines two frameworks developed for this study. First, the CLAUSE framework defines the aspects that needs to be validated: Compliance and Jurisdiction, Legal Competence and Reasoning, Accuracy and Trustworthiness, Understanding Context, Structured Contract Handling, and Ethics, Privacy and Accountability. Second, the LENS-CRAFT rubric provides how to evaluates each generated clause across nine dimensions: Legal Soundness, Expressiveness, Necessity and Completeness, Scenario Alignment, Compliance and Regulatory Alignment, Risk Allocation, Architectural Coherence, Formal Drafting Quality, and Test Robustness.

We focus on four widely-used models that are plausible candidates for integration into legal productivity tools: Claude Haiku 4.5, Gemini 2.5 Flash Lite, GPT 5.4 Nano, and Qwen 3.5 Flash. The experimental design covers 22 clause categories and 34 failure modes (aligned to the CLAUSE framework). The aim is not to find which model is universally best. Instead, we set out to examine which models fail on which kinds of clauses, which failure modes are most severe, and what insights can we derive about holistic evaluation of LLMs in contract drafting.

\paragraph{Contributions.} We bring four contributions in this paper. First, our study presents a clause-level evaluation protocol for LLM contract drafting across 22 legal clause types. Second, our study operationalizes a 34-mode legal failure taxonomy for empirical evaluation. Third, it introduces a risk-sensitive aggregation method, the Max Severity Principle, designed to prevent critical legal defects from being obscured by otherwise high scores. Fourth, it provides a reproducible analysis structure for comparing model behavior across clause categories, failure modes, and legal-risk dimensions.

\section{Related Work and Methodology}
\subsection{Related Work}
\paragraph{Legal LLM evaluation.} LegalBench shifted legal AI evaluation toward task-specific coverage \citep{guha2023legalbench}, while later benchmarks and surveys examine retrieval-augmented legal generation, legal-domain agents, long-form legal QA, and domain-specific NLP limitations \citep{pipitone2024legalbenchrag,li2024legalagentbench,louis2023interpretable,katz2023natural,wang2024legal}. These studies motivate granular evaluations that distinguish legal reasoning, retrieval, factual reliability, and practical legal usability.

\paragraph{Contract-analysis datasets.} Contract-focused benchmarks such as CUAD, LEDGAR, and ContractNLI establish important baselines for clause extraction, contract provision classification, and document-level inference \citep{hendrycks2021cuad,tuggener2020ledgar,koreeda2021contractnli}. LexGLUE also consolidates legal NLU tasks, including contract-related components, into a standardized benchmark suite \citep{chalkidis2022lexglue}. These datasets are complementary to our study; as they evaluate understanding and classification of existing contracts, while our focus is generative clause drafting, severity grading, and failure-mode analysis.

\paragraph{Hallucination, gaps, and legal reliability.} In legal contexts, fabricated reflections or unsupported legal propositions can mislead lawyers, courts, and clients. Work on legal hallucinations, legal-analysis gaps, and human error detection shows that apparently plausible outputs can remain unreliable even when they avoid explicit fake citations \citep{dahl2024large,hou2024gaps,habiblantyer2025phantom,schiller2024human}. Plausibility vs correctness is important for clause drafting because a clause may contain no fabricated authority yet still omit a necessary procedure, exception, or remedy.

\paragraph{Contract drafting and legal language.} Contract generation differs from legal question answering because outputs can be incorporated directly into agreements. Prior work studies LLM-assisted drafting, synthetic contract data, smart legal contracts, expert-annotated contract retrieval, and legal-data-transfer drafting \citep{lam2023contract,kasundra2023contract,chen2023smartcontracts,wang2025acord,thaldar2025datatransfer}. Some of the work on legalese and plain language also exhibits the need for  clarity \citep{martinez2023legalese,schindler2024plain}, but the nuance is that a clearly represented text may still allocate risk improperly or violate regulatory requirements. Legal clause generation evaluation is not semantic syntax problem; but a domain aligned reliability problem. 

\paragraph{Trustworthy and high-risk AI.} Legal drafting is high-risk because users may rely on outputs in binding documents and in some cases may not spot a legal nuance. Research on trustworthy AI, hidden LLM risks, bias, prompt sensitivity, and legal accountability emphasizes robustness, consistency, transparency, and human oversight \citep{liu2022trustworthy,wang2023hidden,garimella2023kelly,qin2024nexus,vargasmurillo2024justice}. Evaluation in high-risk domains must account for asymmetric harms, where one severe error can dominate many minor successes \citep{lawrence2023tightrope}.

\paragraph{Rubric reliability and LLM-as-judge design.} LLM-as-judge evaluation using an LLM coucil, rubric-based scoring, and reliability calibration are foundational for our study \citep{zheng2023judging,liu2023geval,autorubric2026}. Recent studies show that judge choice, prompt design, calibration, and robustness checks can materially affect evaluation reliability \citep{li2025llmjudge}. This motivates our use of explicit scoring anchors, multiple evaluators via LLM council, sampled human review, and Max Severity aggregation rather than relying on a single uncalibrated judge score.

\subsection{Methodology}
\label{sec:methodology}
We adopt a structured empirical benchmarking methodology to evaluate the reliability, robustness, and legal adequacy of LLMs for contractual clause generation. The methodology combines systematically grounded legal evaluation, structured prompt engineering, failure-mode-oriented testing, multi-model comparative benchmarking, and rubric-driven risk assessment.

The study proceeds through a compact eight-stage pipeline: (1) Review legal AI reliability and drafting literature; (2) define the CLAUSE ontology for verifiables; (3) define the LENS-CRAFT rubric for evaluation; (4) select 22 clause categories; (5) generate prompts for B2B, B2C, and B2G by clause, context, failure mode, and legal constraint; (6) run the four models; (7) score normalized outputs through a rubric-driven LLM council; and (8) validate approximately 10\% of outputs through stratified, exception-triggered, and high-severity manual review.

For reproducibility and analysis, the CLAUSE dimensions and LENS-CRAFT dimensions will be open-sourced along with the definitions, representative exemplars, prompt guidance, and model-configuration sheet. The run settings are summarized in \hyperref[app:model-run-settings]{Appendix~A}.

\section{Evaluation Framework}
\subsection{CLAUSE: Prompt Classification and Failure Targeting}
The CLAUSE framework classifies each drafting prompt by the legal function being tested. It is used before generation to ensure that prompts are not treated as generic drafting requests but as structured tests of legal capability. The six CLAUSE dimensions are shown in \cref{tab:clause}.

\begin{table}[t]
\caption{CLAUSE dimensions used for prompt classification.}
\label{tab:clause}
\centering
\small
\begin{tabular}{p{0.09\linewidth}p{0.78\linewidth}}
\toprule
Code & Evaluation target \\
\midrule
C & Compliance and Jurisdiction: regulatory compliance, governing law, venue, and jurisdictional adaptability. \\
L & Legal Competence and Reasoning: doctrine, terminology, enforceability, and contractual logic. \\
A & Accuracy and Trustworthiness: avoidance of hallucination, outdated assumptions, and unsupported legal claims. \\
U & Understanding Context: alignment with party roles, transaction facts, industry, and prompt constraints. \\
S & Structured Contract Handling: definitions, exceptions, cross-references, procedural triggers, and remedies. \\
E & Ethics, Privacy and Accountability: confidentiality, data protection, fairness, and responsible allocation of accountability. \\
\bottomrule
\end{tabular}
\end{table}

\subsection{LENS-CRAFT: Clause Quality and Risk Scoring}
After generation, each clause is evaluated using LENS-CRAFT, a nine-dimensional rubric designed for legal drafting. Each dimension receives a severity score from 1 to 5, where 1 indicates production-ready quality and 5 indicates a highly litigious, void, illegal, or otherwise severe defect. The dimensions are summarized in \cref{tab:lenscraft}.

\begin{table*}[t]
\caption{LENS-CRAFT scoring dimensions. Each dimension is scored on a 1--5 severity scale.}
\label{tab:lenscraft}
\centering
\small
\begin{tabular}{p{0.30\linewidth}p{0.62\linewidth}}
\toprule
Dimension & Evaluation question \\
\midrule
Legal Soundness & Is the provision valid, enforceable, and consistent with active legal constraints? \\
Expressiveness & Is the drafting clear, precise, readable, and free from avoidable ambiguity? \\
Necessity and Completeness & Does the clause include required carve-outs, exclusions, notice mechanics, remedies, and procedural details? \\
Scenario Alignment & Does the clause fit the specified business model, party role, transaction scale, and industry context? \\
Compliance and Regulatory Alignment & Does the clause respect privacy, consumer, security, sectoral, and other regulatory obligations? \\
Risk Allocation & Is liability, indemnity, warranty, and operational exposure allocated in a commercially plausible and legally defensible way? \\
Architectural Coherence & Does the clause remain internally coherent and avoid conflict with definitions, sibling clauses, and boilerplate provisions? \\
Formal Drafting Quality & Does the text follow standard contract-drafting conventions, grammar, and terminology? \\
Test Robustness & Does the output remain stable under prompt variation, edge cases, and adversarial stressors? \\
\bottomrule
\end{tabular}
\end{table*}

\subsection{Risk Scale and Max Severity Principle}
The protocol maps LENS-CRAFT ratings to a five-level risk scale: Level 1, Production-Ready; Level 2, Generally Safe; Level 3, Risk-Prone or Ambiguous; Level 4, Highly Problematic; and Level 5, Highly Litigious. This ordinal mapping follows the broader safety-evaluation principle that deployment decisions should vary with strictness and harm severity rather than average acceptability \citep{flexguard2026}. The final risk level is computed using a Max Severity Principle:
\begin{equation}
R_{final}(x)=\max_{d \in D} s_d(x),
\end{equation}
where $x$ is a generated clause, $D$ is the set of LENS-CRAFT dimensions and critical failure indicators, and $s_d(x)$ is the severity score for dimension $d$. This rule intentionally rejects simple averaging. In legal drafting, a clause that is clear, grammatical, and well structured may still be unacceptable if one dimension reveals a dispositive defect, such as an unlawful data-use authorization or unenforceable liability waiver.

\subsection{Contract nature}
The study evaluates contract clause generation across three transactional contexts: Business-to-Business (B2B), Business-to-Consumer (B2C), and Business-to-Government (B2G). These contexts were selected to capture differing legal, operational, regulatory, and risk-allocation dynamics associated with enterprise negotiations, consumer protection obligations, and public-sector procurement environments.

\section{Dataset and Experimental Design}
\subsection{Clause Categories}
The study evaluates 22 clause categories selected to cover technology contracting, data governance, liability allocation, service operations, and standard boilerplate. \Cref{tab:clauses} groups the clause categories into legal families. These are general contract clauses in AI contracts.

\begin{table*}[t]
\caption{Clause categories evaluated in the study.}
\label{tab:clauses}
\centering
\small
\begin{tabular}{p{0.22\linewidth}p{0.70\linewidth}}
\toprule
Family & Clause categories \\
\midrule
Data, privacy, and security & Confidentiality; Privacy and data protection; Security; Data use; Use of content to train the product. \\
Technology and IP controls & Intellectual property; Restrictions on reverse engineering; Restrictions on use; Advertising or publication rights. \\
Risk allocation and remedies & Indemnity; Warranty; Limitations of liability; Insurance; Safety. \\
Operational service terms & Outage, service interruption, and changes to service; Subcontracting; Third-party involvement; Termination. \\
Dispute and boilerplate terms & Disputes; Jurisdiction; Invalidity or severability; Waiver. \\
\bottomrule
\end{tabular}
\end{table*}

\subsection{Failure Modes}
The evaluation examines 34 failure modes. The failure modes are organized into eight families: legal invalidity, regulatory non-compliance, contextual mismatch, incompleteness, risk-allocation defects, internal inconsistency, drafting ambiguity, and trustworthiness defects. This organization follows the observation that legal AI errors are often not isolated factual mistakes but failures to satisfy the functional requirements of an operative legal text.

\subsection{Models}
The four evaluated systems are Claude Haiku 4.5, Gemini 2.5 Flash Lite, GPT 5.4 Nano, and Qwen 3.5 Flash. They represent a diverse set of widely-used models which are attractive for high-volume legal
drafting workflows. Decoding and evaluation settings were maintained consistently across models where supported to enable comparison. Appendix 1.

\subsection{Prompt and Generation Protocol}
Each input row contains a drafting prompt aligned to a broad CLAUSE category, sub-category, target failure mode and transactional context (B2B, B2C or B2G). This forms a set of static prompts for each such unique situation which can uniformly assess different models. Each prompt is run against each model, producing a model-by-prompt matrix of clause outputs. Outputs are then scored under the LENS-CRAFT rubric and assigned final risk levels under the Max Severity Principle.

\subsection{Risk Scoring using LLM Council and LENS-CRAFT}
Once we obtain a drafted contract clause from a model, an LLM council is employed to determine the best-suited LENS-CRAFT risk levels and reasoning. The LLM council approach uses 3 models - GPT-OSS-120B, Llama-3.3-70B and Gemini 2.5 Flash Lite. Each of these models first scores the drafted clauses using the severity guidance, then each model anonymously ranks each others' responses to find the highest ranked response, which is then chosen as the consensus score of the council.

\subsection{Analysis Approach}
For each drafting condition, the primary outcomes are final overall risk level, risk level for each dimension of LENS-CRAFT, analysis of high-risk rate (Levels 4--5), failure-mode incidence, and clause-category risk distribution. Focus was on distributional measures of score across clauses; LENS-CRAFT; CLAUSE components; and models  rather than only a single aggregate score against each clause. This provides for both - an aggregate risk assessment as well as dimension-wise scores for deeper understanding of exact limitations associated with LLM contract-drafting.

\section{Results}
\label{sec:results}
The evaluation produced 8,956 scored clause outputs (22 clauses x 34 failure modes x 3 transactional contexts minus a few failed generations where Qwen failed to generating certain clauses-prompts, hence those were excluded from the analysis) across each of the four models.

\subsection{Overall Model Comparison}
\Cref{tab:model-results} reports the model-level comparison. GPT-5.4 Nano was the strongest model overall, with the highest Level 1 share and the lowest Level 4--5 share. Claude Haiku 4.5 was comparatively stable but less precise in carve-outs and risk-allocation language. Gemini 2.5 Flash Lite produced fluent clauses but had the highest high-risk share. Qwen 3.5 Flash combined moderate-to-high legal risk with the only large-scale generation failures.

\begin{table}[t]
\caption{Aggregate model-level risk distribution.}
\label{tab:model-results}
\centering
\small
\resizebox{\columnwidth}{!}{%
\begin{tabular}{lccc}
\toprule
Model & Level 1 & Level 4--5 & Summary pattern \\
\midrule
GPT-5.4 Nano & 42.6\% & 12.8\% & Strongest legal robustness \\
Claude Haiku 4.5 & 30.0\% & 25.6\% & Stable but conservative \\
Qwen 3.5 Flash & 13.3\% & 25.5\% & Operationally unstable \\
Gemini 2.5 Flash Lite & 9.3\% & 27.9\% & Fluent but ambiguous \\
\bottomrule
\end{tabular}}
\end{table}

\Cref{tab:risk-distribution} shows the aggregate risk distribution. Level 3 was the dominant category, accounting for 36.3\% of evaluated clauses, followed by Level 4 at 22.9\%. Levels 3 and 4 together represented approximately 59\% of all outputs, indicating that most clauses were not catastrophically defective but still required legal review, redrafting, or substantive correction before deployment.

\begin{table}[t]
\caption{Overall risk distribution across all evaluated clauses.}
\label{tab:risk-distribution}
\centering
\small
\begin{tabular*}{\columnwidth}{@{\extracolsep{\fill}}lp{0.36\columnwidth}rr@{}}
\toprule
Risk level & Meaning & Count & Share \\
\midrule
Level 1 & Production-ready & 2,120 & 23.9\% \\
Level 2 & Generally safe & 1,493 & 16.8\% \\
Level 3 & Ambiguous or risk-prone & 3,222 & 36.3\% \\
Level 4 & Problematic or potentially unenforceable & 2,027 & 22.9\% \\
Level 5 & Highly dangerous or litigious & 7 & $<0.1$\% \\
\bottomrule
\end{tabular*}
\end{table}

The small number of Level 5 outputs suggests that catastrophic legal failures were rare. The much larger Level 3--4 mass is nevertheless central to the study: lightweight models often generated clauses that looked professionally drafted while remaining ambiguous, incomplete, compliance-sensitive, or potentially unenforceable.

\subsection{Transactional Context Effects}
Transactional context materially affected legal quality. \Cref{tab:context-risk-normalized} reports the normalized risk distribution by context, while \cref{tab:b2b-model-results,tab:b2c-model-results,tab:b2g-model-results} report model-level distributions within each context. B2C exhibited the highest concentration of Level 4 risk, suggesting that consumer-facing legal drafting remains the most challenging context for AI models. B2G produced the strongest Level 1 performance overall, indicating that models handled structured public-sector and compliance-oriented drafting more effectively than consumer-oriented negotiations.

\begin{table}[t]
\caption{Normalized risk distribution by transactional context (\%).}
\label{tab:context-risk-normalized}
\centering
\small
\resizebox{\columnwidth}{!}{%
\begin{tabular}{lccccc}
\toprule
Context & L1 & L2 & L3 & L4 & L5 \\
\midrule
B2B & 21.0\% & 17.8\% & 39.2\% & 22.0\% & 0.1\% \\
B2C & 19.0\% & 15.4\% & 38.3\% & 27.2\% & 0.1\% \\
B2G & 31.9\% & 17.4\% & 31.3\% & 19.3\% & 0.1\% \\
\bottomrule
\end{tabular}}
\end{table}

\begin{table*}[t]
\caption{Model-level risk distribution in B2B context.}
\label{tab:b2b-model-results}
\centering
\small
\begin{tabular*}{\textwidth}{@{\extracolsep{\fill}}lccp{0.42\textwidth}@{}}
\toprule
Model & Level 1 & Level 4--5 & Summary pattern \\
\midrule
GPT-5.4 Nano & 35.7\% & 14.7\% & Strongest commercial robustness \\
Claude Haiku 4.5 & 26.1\% & 25.1\% & Stable but generalized drafting \\
Qwen 3.5 Flash & 13.1\% & 22.8\% & Moderate operational instability \\
Gemini 2.5 Flash Lite & 9.2\% & 25.7\% & Fluent but commercially ambiguous \\
\bottomrule
\end{tabular*}
\end{table*}

\begin{table*}[t]
\caption{Model-level risk distribution in B2C context.}
\label{tab:b2c-model-results}
\centering
\small
\begin{tabular*}{\textwidth}{@{\extracolsep{\fill}}lccp{0.42\textwidth}@{}}
\toprule
Model & Level 1 & Level 4--5 & Summary pattern \\
\midrule
GPT-5.4 Nano & 47.9\% & 14.7\% & Strongest consumer-oriented robustness \\
Claude Haiku 4.5 & 24.6\% & 29.9\% & Structurally stable but conservative \\
Qwen 3.5 Flash & 9.0\% & 31.3\% & Highest consumer-context instability \\
Gemini 2.5 Flash Lite & 5.7\% & 32.9\% & Fluent but highly ambiguous under consumer protections \\
\bottomrule
\end{tabular*}
\end{table*}

\begin{table*}[t]
\caption{Model-level risk distribution in B2G context.}
\label{tab:b2g-model-results}
\centering
\small
\begin{tabular*}{\textwidth}{@{\extracolsep{\fill}}lccp{0.42\textwidth}@{}}
\toprule
Model & Level 1 & Level 4--5 & Summary pattern \\
\midrule
GPT-5.4 Nano & 55.5\% & 9.0\% & Strongest public-sector legal robustness \\
Claude Haiku 4.5 & 39.4\% & 21.4\% & Stable and compliance-oriented drafting \\
Qwen 3.5 Flash & 18.4\% & 22.2\% & Moderate public-sector instability \\
Gemini 2.5 Flash Lite & 12.9\% & 24.8\% & Fluent but weak under regulatory complexity \\
\bottomrule
\end{tabular*}
\end{table*}

In B2B settings, the most difficult clauses were intellectual property, indemnity, limitation of liability, outage or service interruption, and termination. These failures primarily reflected commercial risk-allocation instability: weak cap structures, imprecise indemnification carve-outs, and incomplete operational remedies. In B2C settings, the dominant weakness was over-aggressive risk transfer, including broad disclaimers, unilateral modification rights, weak consumer remedies, consent ambiguity in training-data clauses, and jurisdictional assumptions that may conflict with consumer-protection rules. In B2G settings, failures clustered around public-sector liability, government ownership rights, public disclosure exceptions, procurement governance, auditability, and regulatory specificity.

\subsection{Context-Specific Model Trends}
GPT-5.4 Nano performed best overall because it consistently achieved the strongest scores in legal soundness, compliance alignment, risk allocation, and robustness. It generated more operationally usable clauses with stronger procedural completeness and clearer liability balancing. However, failures still emerged in nuanced enterprise edge cases, such as limitation-of-liability language stating that \emph{provider liability shall not exceed fees paid under this agreement} while omitting confidentiality, fraud, and intellectual-property infringement carve-outs, thereby weakening enforceability under enterprise negotiation scenarios.

Claude Haiku 4.5 performed strongly in architectural coherence and formal drafting quality because it produced stable, conservative, and structurally consistent contractual language. Its outputs were generally safer and more predictable, but often overly generalized. For example, language requiring a provider to implement \emph{commercially reasonable safeguards to protect customer information} could omit measurable security standards, breach-notification obligations, and incident-response procedures, reducing operational precision.

Gemini 2.5 Flash Lite demonstrated strong linguistic fluency and readability but underperformed in legal robustness, compliance precision, and prompt stability. It frequently generated contractually plausible but legally soft clauses. For instance, language stating that the \emph{customer agrees that the provider shall not be responsible for any losses arising from service usage} left liability boundaries undefined, ignored statutory limitations, and omitted consumer-protection safeguards, creating significant enforceability and fairness concerns.

Qwen 3.5 Flash showed the weakest operational consistency due to generation instability, higher ambiguity, and weaker procedural integration. Although it generated recognizable legal drafting patterns, it frequently omitted critical legal mechanics. For example, dispute language stating that \emph{disputes shall be resolved appropriately between the parties} omitted governing law, venue, escalation procedures, and arbitration mechanisms, rendering the clause procedurally incomplete and commercially weak.

\subsection{Clause-Level Findings}
For instance, Gemini 2.5 Flash Lite most clearly exhibited this pattern by generating highly readable but legally incomplete clauses, including phrases such as “Provider may process user data as necessary” and “industry-standard security measures,” which lacked procedural safeguards, measurable obligations, and jurisdiction-specific compliance requirements. Qwen 3.5 Flash demonstrated the weakest stability in procedurally complex clauses, producing vague dispute-resolution language such as “Disputes shall be resolved appropriately between the parties,” without specifying governing law, arbitration procedures, or venue requirements. 
The clause-level analysis revealed substantial variation in legal drafting quality across models, particularly in procedurally complex and compliance-sensitive clauses. GPT-5.4 Nano consistently achieved the strongest overall performance through clearer procedural safeguards, liability carve-outs, and operational specificity. Claude Haiku 4.5 produced structurally coherent and conservative drafting but often lacked technical precision. Gemini 2.5 Flash Lite and Qwen 3.5 Flash generated more fluent language yet frequently relied on vague, incomplete, or unenforceable provisions, especially in privacy, liability, security, and intellectual property clauses. Overall, the findings demonstrate that current language models can generate recognizable contractual structures but still exhibit significant weaknesses in enforceability precision, compliance operationalization, and jurisdictionally aware legal reasoning. A full clause-level comparative analysis is provided in \hyperref[app:clause-level-analysis]{Appendix~B}. 

\begin{table}[t]
\caption{Recurring high-risk clause categories and dominant failure patterns.}
\label{tab:difficult-clauses}
\centering
\small
\resizebox{\columnwidth}{!}{%
\begin{tabular}{ll}
\toprule
Clause category & Dominant failure pattern \\
\midrule
Limitation of liability & Overbroad exclusions and weak carve-outs \\
Indemnity & Asymmetrical or incomplete liability allocation \\
Intellectual property & Ownership and license-scope ambiguity \\
Outage/service interruption & Weak remedies and SLA specificity \\
Jurisdiction & Cross-border enforceability assumptions \\
Privacy and data protection & Compliance and consent incompleteness \\
\bottomrule
\end{tabular}}
\end{table}

\subsection{LENS-CRAFT and CLAUSE Findings}
Across LENS-CRAFT dimensions, models were strongest in Scenario Alignment, Architectural Coherence, and Formal Drafting Quality, but weakest in Test Robustness, Compliance, Risk Allocation, and Expressiveness. GPT-5.4 Nano had the strongest average LENS-CRAFT profile, while Gemini's weakest scores were Test Robustness (2.60), Compliance (2.53), Expressiveness (2.44), Necessity and Completeness (2.44), and Risk Allocation (2.31). These results show that Gemini optimized fluency more than enforceability rigor, while GPT-5.4 Nano maintained the best balance between drafting precision, contextual alignment, and legal robustness.

The CLAUSE component and sub-component risk analysis is provided in \hyperref[app:clause-component-analysis]{Appendix~C}. Compliance and Jurisdictional Adaptability was the highest-risk component, driven by weak localization to GDPR, consumer protection, export-control, procurement, governing-law, and venue constraints. Legal Competence and Reasoning failures involved carve-outs, indemnity triggers, waiver mechanics, survivability, causation, proportionality, and fault allocation. Structured Contract Elements Handling captured missing notices, escalation procedures, survival clauses, fallback remedies, and operational contingencies. Understanding Context was comparatively stronger, but models remained unstable under incomplete, multi-party, hybrid B2B/B2G, and use-restriction prompts. Ethics, Privacy, and Accountability was strongest numerically, likely because models have substantial exposure to standardized privacy and accountability language.
\begin{table}[t]
\caption{Most fragile CLAUSE sub-components overall.}
\label{tab:fragile-clause-subcomponents}
\centering
\small
\begin{tabular}{p{0.72\columnwidth}c}
\toprule
Sub-component & Risk \\
\midrule
Legal Standards & 3.43 \\
Nuance and Interpretation & 3.35 \\
Nuance Omission & 3.33 \\
Jurisdiction & 3.23 \\
Ambiguity Handling & 3.18 \\
Accuracy Rate & 3.10 \\
Prompt Sensitivity & 3.06 \\
\bottomrule
\end{tabular}
\end{table}

Across CLAUSE components, lightweight models failed most severely when legal nuance, cross-jurisdiction adaptation, procedural completeness, and adversarial robustness had to operate simultaneously. They succeeded most strongly when drafting patterns were standardized, repetitive, stylistically predictable, and heavily represented in training data.

\subsection{Max Severity Ablation}
The empirical distribution supports the Max Severity Principle. Because the strongest model dimensions were often formal, structural, or contextual, average-dimensional scoring would tend to reward fluency even when a clause contained a legally decisive defect. The concentration of outputs in Levels 3 and 4 shows why a severity-aware rule is necessary: a single missing consumer-law safeguard, liability carve-out, compelled-disclosure exception, or jurisdictional limitation can dominate otherwise polished drafting.

\section{Discussion}
\label{sec:discussion}
Our analysis reflects three key aspects. (1) Model performance in legal drafting is clause-specific and context-sensitive. A model that produces fluent boilerplate may still fail on regulatory, consumer-facing, or risk-allocation clauses. (2) )he findings show a clear hierarchy among the tested lightweight models: GPT-5.4 Nano was strongest, Claude Haiku 4.5 was moderate and structurally stable, Gemini 2.5 Flash Lite was moderate but risk-prone, and Qwen 3.5 Flash was the weakest and most operationally unstable. (3))Legal defects (including severe ones) can be hidden by average scoring because formal drafting quality and expressiveness may remain high even when legal soundness or compliance fails. This is a critical factor given the emerging challenge of automation bias in high skilled and sensitive domains.  

The central empirical pattern is that lightweight LLMs are substantially more capable of reproducing the stylistic and structural characteristics of contractual drafting than consistently operationalizing legally robust, enforceable, and compliance-sensitive provisions. Across models, fluency outperformed robustness, regulatory precision, and risk-allocation reliability. This gap was most visible in B2C clauses, where consumer protection, fairness, disclosure, and statutory override concerns produced the highest concentration of legally risky outputs.

The evaluation also illustrates practical governance lessons. Structured rubrics are necessary for reproducibility; clause-specific reporting helps identify recurring vulnerabilities; and longitudinal comparisons can track whether model updates improve or degrade legal reliability. However, automated evaluation should not replace legal review. For Level 3--5 outputs, human expert review remains necessary before any clause is used in a live agreement. We believe that such an assessment can help classify the clauses for human attention, thereby becoming a reliability layer in the solution. 

\section{Limitations and Way Forward}
This study has six limitations. (1) Our study is not anchored to a single jurisdiction or doctrinal system; statutory interpretation, procedural requirements, drafting conventions, evidentiary standards, enforceability thresholds, and legal terminology may vary even across common-law systems. (2) The regulatory analysis is intentionally broad rather than sector-specific: the study does not operationalize detailed rules for financial services, healthcare, procurement, telecommunications, defense, or critical infrastructure. (3) The clause set focuses on recurring AI and technology-contract provisions, so the findings should not be generalized to specialized instruments such as pharmaceutical licenses, derivatives, defense procurement, energy contracts, clinical-trial agreements, or highly customized enterprise procurement frameworks. (4) The evaluated systems are general-purpose models rather than specialized legal models. The results therefore characterize lightweight models operating in legal drafting contexts, not purpose-built legal AI systems. (5) CLAUSE and LENS-CRAFT are principle-based frameworks, and legal practitioners may reasonably disagree about enforceability, proportionality, drafting sufficiency, or commercial reasonableness. (6) The LLM council and sampled human review cannot fully reproduce real-world negotiation, litigation exposure, judicial discretion, evolving regulation, organizational risk tolerance, or factual disputes. The study is thus an empirical benchmarking exercise, not a definitive legal-validity assessment.

Future work should develop sector-specific and jurisdictionally grounded benchmarks, operational legal-risk simulations, cost-risk comparisons, inter-rater agreement studies, and longitudinal robustness tests under changing regulatory conditions. Continuous evaluation pipelines are needed to detect emerging drafting weaknesses, regulatory misalignment, liability-allocation failures, and adversarial robustness degradation in production legal drafting workflows.

\section{Ethical and Governance Considerations}
Legal drafting systems raise confidentiality, privacy, and professional-responsibility concerns. If prompts or source contracts contain sensitive information, third-party API use may create disclosure risks. The study therefore discloses whether prompts are synthetic, public-template-derived, or adapted from confidential materials, and it treats all generated outputs as research artifacts rather than deployable legal advice.

There is also a risk of overreliance. LLM-generated clauses can appear authoritative even when legally defective. Presenting risk levels, rationales, and failure modes can reduce unwarranted trust by showing users why review is required. Finally, the study reports model weaknesses responsibly, emphasizing defensive evaluation and legal oversight rather than adversarial exploitation.

\section{Conclusion}
This paper presents a risk-sensitive empirical evaluation of LLM-generated contract clauses. Across 8,869 evaluated outputs, lightweight models demonstrated substantial fluency in contractual language generation but persistent deficiencies in enforceability, completeness, compliance, and risk allocation. GPT-5.4 Nano exhibited comparatively stronger legal robustness, whereas Gemini and Qwen displayed elevated concentrations of Level 4 risk outcomes, particularly in regulatory and operational drafting scenarios. Although GPT-5.4 Nano demonstrated comparatively stronger performance across multiple LENS-CRAFT dimensions, including legal soundness, compliance alignment, risk allocation, and robustness, these findings should not be interpreted as establishing the model as a definitive gold standard for AI-assisted legal drafting. Contractual interpretation remains highly dependent on downstream operational realities, jurisdictional nuance, transactional context, sector-specific obligations, negotiation dynamics, and factual circumstances that cannot be fully captured through generalized benchmarking alone. Consequently, the practical reliability and legal adequacy of generated clauses can only be meaningfully assessed through domain-specific downstream evaluations conducted within the intended context of use.
 
\bibliography{main}
\bibliographystyle{icml2026}

\clearpage
\onecolumn
\appendix
\section*{Appendices}
\addcontentsline{toc}{section}{Appendices}
\section{Model Run Settings}
\label{app:model-run-settings}
All model runs used consistent settings within each stage of the pipeline. For clause generation, the temperature was set to 0.2, the maximum output length was set to 2,000 tokens, and no additional formatting constraints were applied beyond the instruction to return only the clause text. For LENS-CRAFT evaluation, council models used a temperature range of 0.1--0.2 and a maximum output length of 3,000 tokens for each individual evaluation. The evaluation prompt provided guidance on response structure, but structured output was not enforced because not all models available through OpenRouter supported structured output functionality. After the council responses were ranked, Gemini 2.5 Flash Lite was used only to restructure the highest-ranked response into the available structured-output schema.

\section{Clause-Level Comparative Analysis}
\label{app:clause-level-analysis}
\begin{table}[!htbp]
\caption{Clause-level comparative analysis of dominant legal-risk patterns across evaluated language models.}
\label{tab:clause_level_analysis}
\centering
\scriptsize
\resizebox{\textwidth}{!}{%
\begin{tabular}{p{3.0cm} p{3.2cm} p{4.8cm} p{4.8cm}}
\toprule
\textbf{Clause Category} & \textbf{Primary Weakness} & \textbf{Dominant Failure Pattern} & \textbf{Example Problematic Language} \\
\midrule

Confidentiality Clauses &
Ambiguity and weak procedural safeguards &
Missing survival clauses, vague disclosure obligations, and incomplete incident procedures &
``reasonable efforts''; ``appropriate safeguards'' \\

Privacy \& Data Protection Clauses &
Compliance incompleteness &
Missing lawful basis requirements, weak retention obligations, and inadequate cross-border safeguards &
``Provider may process user data as necessary.'' \\

Security Clauses &
Operational vagueness &
Undefined security standards, weak breach notification procedures, and missing escalation mechanisms &
``industry-standard security measures'' \\

Safety Clauses &
Lack of operational accountability &
Missing audit obligations, escalation procedures, and responsibility allocation &
Generalized safety commitments without enforcement structures \\

Intellectual Property Clauses &
Ownership ambiguity &
Conflicting derivative-work provisions, overlapping rights structures, and licensing inconsistencies &
Contradictory jointly developed IP ownership provisions \\

Use of Content to Train Product Clauses &
Consent and governance failures &
Failure to distinguish analytics usage, model training, and commercialization rights &
``Provider may use submitted content to improve services.'' \\

Indemnity Clauses &
Imbalanced liability allocation &
Missing defense-control rights, vague causation standards, and incomplete notice procedures &
``Customer agrees to indemnify Provider for claims arising from use.'' \\

Warranty Clauses &
Overbroad disclaimers &
Undefined performance standards and insufficient consumer-law carve-outs &
``Services are provided as-is.'' \\

Limitation of Liability Clauses &
Extreme enforceability risk &
Absolute liability exclusions, omission of fraud carve-outs, and disproportionate risk allocation &
``Provider shall not be liable for any damages under any circumstances.'' \\

Advertising \& Publication Rights Clauses &
Consent and reputational-risk failures &
Broad publication permissions without revocation rights or consent boundaries &
Broad publication rights without revocation procedures \\

Invalidity \& Severability Clauses &
Doctrinal ambiguity &
Weak jurisdictional logic and unenforceable interpretive standards &
``informed by the parties' good faith reliance on their respective internal knowledge bases'' \\

Waiver Clauses &
Procedural invalidity &
Blanket waivers and unconstrained relinquishment language &
``User waives all legal claims permanently.'' \\

Subcontracting Clauses &
Accountability discontinuity &
Missing audit rights, liability continuity, and approval requirements &
``Provider may subcontract services at its discretion.'' \\

Third-Party Involvement Clauses &
Flow-down compliance ambiguity &
Missing downstream compliance obligations and confidentiality continuity &
Broad third-party access without audit rights \\

Dispute Resolution Clauses &
Procedural incompleteness &
Venue ambiguity, missing escalation sequencing, and absent arbitration frameworks &
``Disputes shall be resolved appropriately between the parties.'' \\

Insurance Clauses &
Undefined operational thresholds &
Missing minimum coverage requirements and proof-of-insurance obligations &
Generic insurance requirements without measurable thresholds \\

Outage \& Service Interruption Clauses &
SLA ambiguity &
Weak remedies, unilateral modification powers, and inadequate customer protections &
``Provider may modify services at any time.'' \\

Jurisdiction Clauses &
Governing-law incompleteness &
Missing venue specification and conflict-of-law provisions &
``This Agreement shall be governed by applicable law.'' \\

Termination Clauses &
Procedural imbalance &
Missing cure periods, post-termination obligations, and survival provisions &
Broad unilateral termination rights \\

Restrictions on Reverse Engineering Clauses &
Overbreadth and enforceability conflicts &
Missing interoperability exceptions and lawful research allowances &
Absolute prohibitions on reverse engineering \\

Data Use Clauses &
Ownership and reuse ambiguity &
Failure to distinguish analytics usage, commercialization, and training rights &
Broad reuse rights without accountability safeguards \\

Restrictions on Use Clauses &
Vagueness and operational ambiguity &
Undefined prohibited-use categories and absent enforcement mechanisms &
``Users shall not misuse the services.'' \\

\bottomrule
\end{tabular}}
\end{table}

\clearpage
\section{CLAUSE Component Risk Analysis}
\label{app:clause-component-analysis}
\begin{table}[!htbp]
\caption{CLAUSE component and sub-component risk analysis.}
\label{tab:clause-component-summary}
\centering
\small
\resizebox{\textwidth}{!}{%
\begin{tabular}{p{0.20\linewidth}cp{0.27\linewidth}p{0.33\linewidth}}
\toprule
CLAUSE component & Avg. risk & Most fragile sub-components & Dominant failure pattern \\
\midrule
Compliance and Jurisdictional Adaptability & 3.04 & Legal Standards (3.43); Jurisdiction (3.23); Regulatory Compliance (3.08) & Cross-jurisdiction enforceability failures, governing-law ambiguity, missing statutory safeguards, vague regulatory language. \\
Legal Competence and Reasoning & 2.66 & Nuance and Interpretation (3.35); Time-Series Structuring (2.83); Legal Principle Application (2.77) & Weak handling of doctrinal nuance, temporal obligations, proportionality, survivability, causation, and effort standards. \\
Structured Contract Elements Handling & 2.59 & Nuance Omission (3.33) & Missing notice obligations, escalation procedures, survival clauses, exception handling, and fallback remedies. \\
Understanding Context & 2.60 & Ambiguity Handling (3.18) & Context-sensitive drafting instability under incomplete, multi-party, or hybrid transactional prompts. \\
Accuracy, Reliability, and Trustworthiness & 2.57 & Accuracy Rate (3.10); Prompt Sensitivity (3.06); Parametric Knowledge Reliance (2.55) & Legally plausible but unreliable drafting, unsupported assumptions, and instability under perturbation. \\
Ethics, Privacy, and Accountability & 2.01 & Accountability Allocation; Data Governance; Privacy Precision & Broad data rights, weak governance assignment, and incomplete privacy safeguards. \\
\bottomrule
\end{tabular}}
\end{table}

\end{document}